\documentclass[journal]{IEEEtran}
\usepackage{multirow}
\usepackage{amsmath}
\usepackage{booktabs}

\ifCLASSINFOpdf
   \usepackage[pdftex]{graphicx}
\else
\fi
\begin{document}
\title{Signature-Graph Networks}

\author{Ali~Hamdi,
        Flora~Salim,
        Du Yong~Kim,
        and~Xiaojun~Chang
\thanks{Ali Hamdi, Flora Salim and Xiaojun Chang are with the School of Computing Technologies, RMIT University, Australia, ali.ali@rmit.edu.au, flora.salim@rmit.edu.au, xiaojun.chang@rmit.edu.au.}% <-this % stops a space
\thanks{Du Yong Kim is with the School of Engineering, RMIT University, Australia, duyong.kim@rmit.edu.au.}% <-this % stops a space
\thanks{Manuscript received October 14, 2021; revised 15 January 2022.}}


% \markboth{IEEE TNNLS, Special Issue on Deep Neural Networks for Graphs: Theory, Models, Algorithms and Applications}%
% \markboth{IEEE Transactions on Neural Networks and Learning Systems}
\markboth{Ali Hamdi \MakeLowercase{\textit{et al.}}: Signature-Graph Networks}
{Ali Hamdi \MakeLowercase{\textit{et al.}}: Signature-Graph Networks}



\maketitle

\begin{abstract}
We propose a novel approach for visual representation learning called Signature-Graph Neural Networks (SGN). SGN learns latent global structures that augment the feature representation of Convolutional Neural Networks (CNN). SGN constructs unique undirected graphs for each image based on the CNN feature maps. The feature maps are partitioned into a set of equal and non-overlapping patches. The graph nodes are located on high-contrast sharp convolution features with the local maxima or minima in these patches. The node embeddings are aggregated through novel Signature-Graphs based on horizontal and vertical edge connections. The representation vectors are then computed based on the spectral Laplacian eigenvalues of the graphs. SGN outperforms existing methods of recent graph convolutional networks, generative adversarial networks, and auto-encoders with image classification accuracy of $99.65$\% on ASIRRA, $99.91$\% on MNIST, $98.55$\% on Fashion-MNIST, $96.18$\% on CIFAR-10, $84.71\%$ on CIFAR-100, $94.36$\% on STL-10, and $95.86$\% on SVHN  datasets. We also introduce a novel implementation of the state-of-the-art multi-head attention (MHA) on top of the proposed SGN. Adding SGN to MHA improved the image classification accuracy from $86.92$\% to $94.36$\% on the STL10 dataset.
\end{abstract}


% \fi
\IEEEpeerreviewmaketitle

\section{Introduction}
\IEEEPARstart{I}{n} visual representation learning, Convolutional Neural Networks (CNN) have been widely utilised to extract effective feature representations from images. However, CNNs ignore useful global information due to the isotropic nature of their receptive fields \cite{luo2016understanding} and, thus, CNNs are affected by challenging foreground and backgrounds noise. Recent CNN architectures have large sets of layers to adapt to the increasing size and complexity of training data \cite{hamdi2020flexgrid2vec}. Such complex and deep models usually suffer from overfitting when trained on relatively small data. Recently, the overfitting problem has been addressed by different techniques such as data augmentation \cite{masi2016we}. However, traditional augmentation techniques are tackled by visual uncertainties in colours and shapes in images. 
% \textcolor{red}
{For example, augmenting an image of a bird by changing colour may result in a bird of different group, or, rotating an image of a flower has no effect.} To overcome these challenges, images can be represented as graphs to denote global features and integrated into CNN. Yet, traditional CNN cannot deal with such graphs of irregular structure. 
% \textcolor{red}{We propose a graph-based end-to-end trainable network that learns unique local and global visual embedding.}
These CNN limitations can be solved using recent advances in graph neural networks or graph based methods.

There are various graph representations such as region-based \cite{li2018beyond}, part-based \cite{gao2019graph}, and predefined skeleton \cite{yan2018spatial}. 
% \textcolor{red}
{Graph-based methods improved CNN performance as they capture important global features embedding to produce accurate image representations \cite{li2018beyond}.} However, these graph architectures might misrepresent significant local structures and need high computational resources \cite{hamdi2020flexgrid2vec}. Besides, graph models were mainly developed to represent multiple objects in an image for tasks such as detection and tracking. There is less effort in representing the whole image as a feature vector based on graphs that the proposed SGN is addressing. 
% \textcolor{red}
{These graph-designs have been accompanied by complex models, such as context-aware, attentional and memory-based, to enhance visual representation learning \cite{Chen_2019_CVPR,Si_2019_CVPR,Wang_2019_CVPR,Zhao_2019_CVPR,shen2018person}.} 
% \textcolor{red}
{In contrast to these complex architectures, we propose to learn visual feature embeddings through straightforward and accurate signature-like graphs.} We reinvent the idea of capturing global attentions to support the CNN extracted feature embeddings.

\begin{figure*}[!ht]
\begin{center}
   \includegraphics[width=0.99\linewidth]{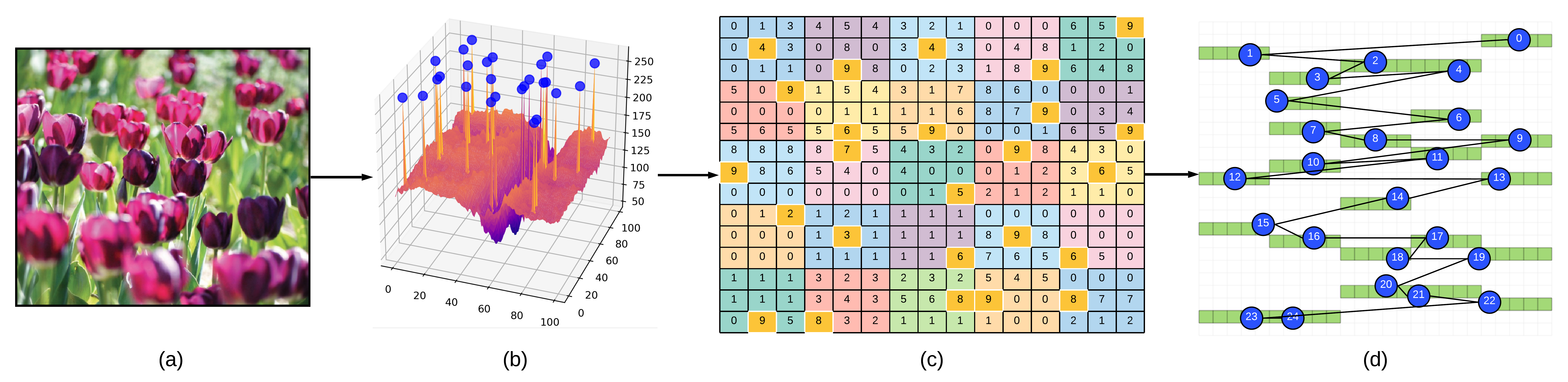}
\end{center}
   \caption{
%   \textcolor{red}
   {(a) heat-map of an image feature map},
   (b) the heat-map with selected coordinates of  convolution features with local maxima in different patches.
%   \textcolor{red}
   {(c and d) constructing Signature-Graphs of the selected nodes' coordinates to aggregate the feature embeddings.}}
   \label{signature-graphs}
\end{figure*}

Signature-Graphs tend to have unique and similar representations for images from the same class and different from the others. We extract the CNN feature maps; then we partition them into a set of equal and non-overlapping patches. Each patch is represented by a node in the Signature-Graph. Fig. \ref{signature-graphs} visualises the construction of the Signature-Graphs. 
% \textcolor{red}
{It shows a heat-map of an image feature map form which the nodes are located to connect the Signature-Graphs. These nodes represent sharp high-contrast features convolutional features with the local maxima in their respective patches.} Similarly, the graph could be constructed with nodes of the local minima of different patches. 
% \textcolor{red}
{To connect these nodes, two novel Signature-Graphs designs are proposed based on horizontal and vertical edge-connections. SGN offers a straightforward yet accurate method to learn useful visual features.} It can be easily extended with other representation learning techniques. For example, multi-head attention (MHA) networks can be employed on top of the SGN to learn more useful features using attention mechanism. Other CNN baseline networks can also be combined with SGN. The experimental results show that SGN has significantly improved the accuracy of these methods.

In this paper, we present a set of contributions to the graph design, CNN vector augmentation, and global attentional feature computing computing, as follows:
\begin{itemize}
    \setlength\itemsep{1mm}
    \item 
    % \textcolor{red}
    {The proposed SGN augments CNN local features with graph-based global feature embeddings. Particularly, we introduce a new feature embedding method, Signature-Graphs, that capture unique global attentions via connections among different image regions.}
    \item We implement a novel design architecture of SGN with the state-of-the-art MHA mechanism. Specifically, we feed the MHA with SGN output vectors as key vectors that capture more accurate attentions than using the existing method.
    \item We evaluate the proposed SGN on seven benchmark multi-class and binary image classification datasets. SGN achieves superior performance over existing methods.
\end{itemize}%\vspace{-5mm}

\begin{figure*}[!ht]
\begin{center}
   \includegraphics[width=0.99\linewidth]{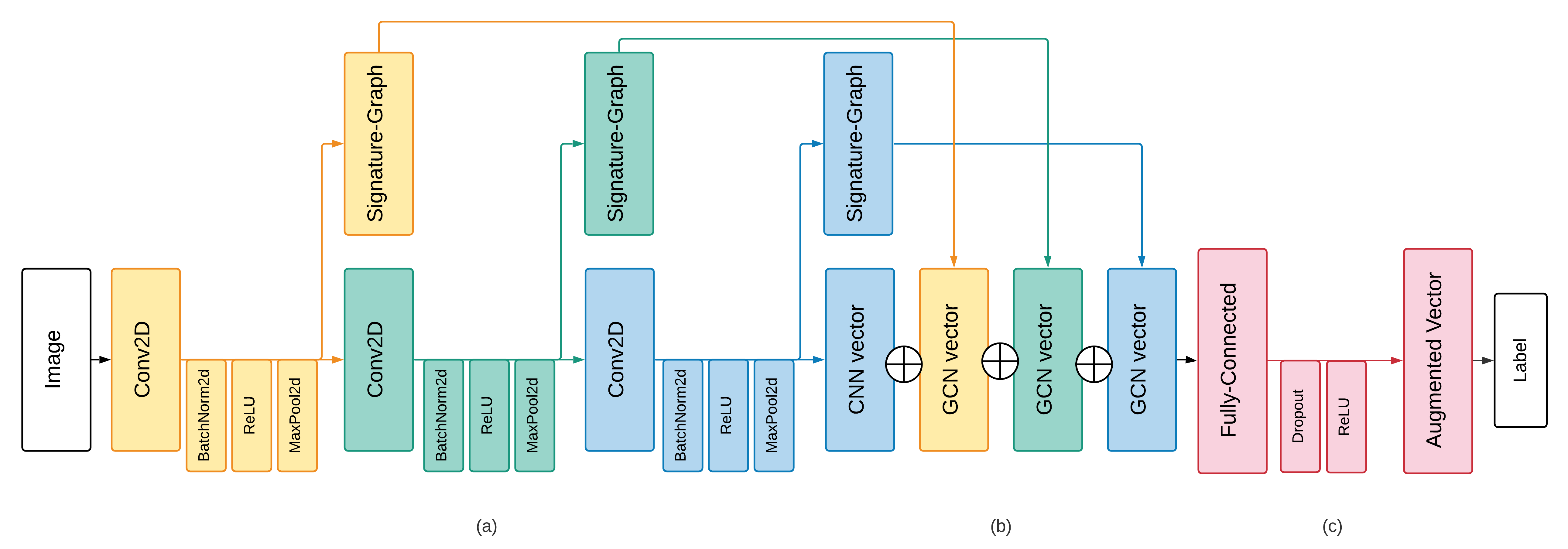}
\end{center}
   \caption{SGN consists of three convolutional blocks (a). Each block is followed by batch normalisation, Relu activation, and max-pooling layers. The feature maps of each convolutional block are injected in the subsequent convolutional block and a Signature-Graph layer. The Signature-Graph layer computes the global features via a Signature-Graphs, as in Fig. \ref{signature-graphs}. The output of the last convolutional block is concatenated with the output of the three Signature-Graph layers (b). A fully-connected layer is attached at the end of SGN to produce the final augmented feature vector.}%\vspace{-5mm}
   \label{VAGCN_2}
\end{figure*}

\section{Related Work}%\vspace{-3mm}
Graph-based models have been widely used in multiple research tasks, such as multi-label recognition \cite{Chen_2019_CVPR}, skeleton-based action recognition \cite{Si_2019_CVPR}, point-cloud semantic segmentation \cite{Wang_2019_CVPR}, large-scale object detection \cite{Xu_2019_CVPR}, 3D pose regression \cite{Zhao_2019_CVPR}, and person re-identification \cite{shen2018person}. Recently, graph-based models have been accompanied by context-aware, attentive and memory-based models.
Chen et al., \cite{Chen_2019_CVPR} utilised graph convolutional network (GCN) for multi-label image recognition. Si et al., \cite{Si_2019_CVPR} proposed an attention-based LSTM with GCN for skeleton-based action recognition. GCN with attention model has also developed in \cite{Wang_2019_CVPR} for point cloud semantic segmentation. Xu et al., \cite{Xu_2019_CVPR} presented a spatial-aware graph relation network for object detection in large-scale data. 
% Zhao et al., \cite{Zhao_2019_CVPR}.
These models capture important global features that support accurate image representations. The attention and memory weights are updated throughout the learning process. 
% \textcolor{red}
{In contrast to the previous deep and complex architectures, we propose to learn visual embeddings via Signature-Graphs that are uncomplicated yet accurate compared to state-of-the-art visual representation methods. }We present a simple implementation to learning important global features in comparison attention and memory-based complex approaches.

There are multiple image graph-based representations. These models try to represent the image in different ways, such as region-based \cite{li2018beyond}, part-based \cite{gao2019graph}, and predefined skeleton \cite{yan2018spatial}. 
These approaches outperform CNN in encoding long relations between image regions \cite{li2018beyond,Si_2019_CVPR}.
Motivated by these approaches we propose a new graph representation method, the so-called, Signature-Graphs. The designed Signature-Graphs tend to overcome the limitations of these previous graph architectures, such as losing significant local structures and the need for high computation resources. 
% \textcolor{red}
{The Signature-Graphs tend to offer robust visual feature embeddings via undirected-graph design.}
Previous graph-based models were developed for visual tasks with multiple objects in an image such as object detection and tracking. However, there is less effort in the case of dealing with the whole image with no predefined semantic labels. The proposed SGN solves this issue by augmenting the CNN vectors with the new Signature-Graphs feature vectors.

\section{Signature-Graph Neural Networks}
The proposed SGN is designed as a mechanism to augment the CNN feature vectors with graph-based global information. Fig. \ref{VAGCN_2} shows the proposed Signature-Graphs. Specifically, we propose to attach a Signature-Graph layer to each convolutional block. The Signature-Graph layer tends to extract global features that are missed in the CNN classical design of the receptive fields. 

\subsection{Convolutional Feature Extraction}
{SGN is a modified architecture of CNN. SGN starts with the extraction of the convolutional feature extraction. Typically, a CNN is composed of a series of convolutional and pooling layers to extract the significant features in an image $I$. The image $I$ is characterised by $I_h$ height, $I_w$ width, and $I_c$ colour channels. The CNN convolutional product is computed as a 2-dimensional matrix that contains the sum of the element-wise multiplication of the image $I$ and a convolutional filter $F$ that must have the $F_c$ equals to the $I_c$.}%\vspace{-5mm} %what does the last sentence mean?

\begin{equation}\label{$Conv2D$}
    Conv2D(I,F)_{x,y} = \sum^h_{i=1} \sum^w_{j=1} \sum^c_{k=1} K_{i,j,k} I_{x+i-1,j+j-1,k}
\end{equation}
where, $I$ and $F$ are the image and convolutional filter, $x$ and $y$ denote the image coordinates, and $h$, $w$, and $c$ are the image attributes. The $Conv2D$ in Eq. \ref{$Conv2D$} learns a feature map of the image $I$.
The output feature maps are then passed to a pooling layer to down-sample the image features. Pooling is using a sliding window, similar to the convolution, although it keeps the max or average of the pixels values instead of the multiplication.

{CNNs are affected by the overfitting problem. This happens when a CNN based method extensively learns the convolutional features through a very complex architecture. In such a scenario, the learnt model performs poorly in the testing on unseen images. Therefore, we utilised a batch normalisation layer before the max-pooling layer to standardise the input pixel values. This normalisation process makes the feature maps to have a mean of zero and a standard deviation of one. This layer observes the learning statistics and updates the standardised data. We also add a non-linear Rectified Linear Unit (Relu) activation layer before the pooling process.} The CNN final feature maps are flattened in a 1-dimensional vector and connected to a fully-connected layer. However, it is not practical to connect all the neurons with all possible region of the input features. This would make the process too complex with many weights to train. Therefore, CNNs depend on a receptive field \cite{luo2016understanding}. The receptive field is simply a 2-dimensional kernel which contains the pixels that are fully-connected. Although this process enhances the convolutional feature learning, it limits the CNN to learn local features from different image regions. Therefore, in this paper, we propose to augment the CNN output feature vector with the global image feature through novel \textit{Signature-Graph}.

\subsection{Signature-Graphs}
We propose to construct a unique graph from each feature map. In the following sub-sections, we explain three main components of the Signature-Graphs as follows:
\begin{itemize}
    \item \textit{Signature-Graph construction} 
    % \textcolor{red}
    {sub-section discusses the selection of the nodes' coordinates and their correspondent visual features.}
    \item \textit{Spectral graph computing} 
    % \textcolor{red}
    {sub-section introduces the computing of the Eigenvalues over the Laplacian matrix to normalise the node feature vector embeddings.}
    \item \textit{Multi-Head Attention} sub-section defines the employed multi-head attention mechanism and how the SGN contributes to it.
\end{itemize}

\subsubsection{Signature-Graphs Construction}
The output standardised and pooled convolutional feature maps are injected into a Signature-Graph layer. The Signature-Graph layer applies graph spectral theory over the convolutional maps. In order to implement the spectral graph methods, we first construct a Signature-Graph. The feature map is divided into a set of equal and non-overlapping patches.

Each patch is a 2-dimensional array of convolutional features. Each patch is represented by one node in the graph. To select the node coordinates in the patch, we consider the feature with the local maxima and minima. The hypothesis behind these selections is that the local maxima and minima of the convolutional feature represents the highest-contrast points in the respective patches. 
% \textcolor{red}
{Therefore, we suppose that positioning the graph nodes at these locations can offer effective representations due to the expected sharpness of these points.}
So, the Signature-Graph nodes can be,
\begin{equation}
    V, X = \max(p_{f}) \rightarrow \max(p_{f})_{row}; \forall p \in P
\end{equation}
or,
\begin{equation}
    V, X = \min(p_{f}) \rightarrow \min(p_{f})_{row}; \forall p \in P
\end{equation}
where $P$ is a set of patches of the feature map, $\max(p_{f})$ is the feature with the highest value in a patch, and $\min(p_{f})$ refers to the feature that has the lowest value in a patch. The output $V$ and $X$ are the nodes and node attributes. The $V$ nodes are computed as a set of CNN feature coordinates in the given patch, and their corresponding rows are employed as the node attributes $X$, as visualised in Fig. \ref{signature-graphs} (d). We utilised the computed $V$ and their attributes $X$ to construct the Signature-Graph, as in Eq. \ref{G}. 

\begin{equation}\label{G}
    G = (V, E)
\end{equation}
where $G$ is the Signature-Graph, $V$ the graph nodes, and $E$ the graph edges that are computed based on the distances between the nodes' coordinates. $G$ is an undirected graph. 

We propose a new graph design to extract the most useful features form the above-explained $G$. We propose to connect the graph nodes through horizontal or vertical edges passing across the image patches. The idea of connecting horizontal and vertical graph components is presented in multiple research studies, such as squaring a square \cite{stewart1997squaring} and planar graphs \cite{chaplick2012planar}. We assume that SGN with horizontal and vertical Signature-Graphs could capture unique latent connections between different local regions in an image. Later, in this paper, the experimental results show that SGN with horizontal and vertical edges outperform each other in different datasets.

\subsubsection{Spectral Graph Computing}
The constructed Signature-Graph has a set of spectral characteristics that can be computed through different matrices such as the degree matrix $D$, adjacency matrix $A$, and incidence matrix $M$. 
The degree matrix $D$ is a diagonal matrix that contains the degree of each node in the graph.
The adjacency matrix $A$, or connection matrix, has boolean representations of the node positions. 
The incidence matrix $M$ represents the connections between the graph nodes and edges.
The Laplacian matrix $L$ is defined as:
\begin{equation}
    \mathcal{L}_{n \times n} = D - A
\end{equation}
where $n$ denotes the number of nodes in the Signature-Graph. The Laplacian $\mathcal{L}$ can be normalised as a Kirchhoff matrix, as follows:
\begin{equation}\label{L2}
    \mathcal{L}_{i j}(G)=\left\{\begin{array}{ll}
    1 & \text{if } i=j \text{ and } d_{j} \neq 0 \\
    -\frac{1}{\sqrt{d_{i} d_{j}}} & \text{if } i \text{ and } j \text{ are adjacent} \\
    0 & \text{otherwise}
    \end{array}\right.
\end{equation}
where $d$ is an element of the degree matrix $D$.

The Laplacian matrix satisfies $\mathcal{L} = M^T M$ with $M$ is the incidence matrix, and $M^T$ is its transpose. Elements in $M$ are described as $M_{ev}$ with $e$ connects nodes $i$ and $j$ and $v$ node as in Eq. \ref{Mev}. %I do not understand what it means.

\begin{equation}\label{Mev}
M_{e v}=\left\{\begin{aligned}
1, & \text { if } v=i \\
-1, & \text { if } v=j \\
0, & \text { otherwise }
\end{aligned}\right.
\end{equation}

The eigendecomposition of $\mathcal{L}$ with eigenvectors and eigenvalues $\lambda$ can be computed as in Eq. \ref{lamda2}

\begin{equation}\label{lamda2}
\begin{aligned}
\lambda_{i} &=\mathbf{v}_{i}^{\top} L \mathbf{v}_{i} \\
&=\mathbf{v}_{i}^{\top} M^{\top} M \mathbf{v}_{i} \\
&=\left(M \mathbf{v}_{i}\right)^{\top}\left(M \mathbf{v}_{i}\right).
\end{aligned}
\end{equation}

We normalise the node embeddings with the eigenvalues of the graph based on a Fourier basis. Fourier basis eigenvectors and their corresponding eigenvalues $\Lambda$ represent the direction and variance of the graph Laplacian. The decomposed eigenvectors are computed as a matrix $U \in R^{n \times n}$ that contains $\eta_{n}$ where $n$ dimension is the same as node counts. These eigenvectors have a natural signal-frequency interpretation for the graph. The spectral Fourier basis decomposition produces the matrix \textit{U} that diagonalises the Laplacian as in Eq. \ref{L}, where $\Lambda$ is a diagonal matrix of non-negative real eigenvalues, see Eq. \ref{lamda}, $A$ is the adjacency matrix, and $I$ is the input image. 

\begin{equation}\label{L}
    \mathcal{L} = diag (A + I)^{-1} = U \lambda U^{T}
\end{equation}

\begin{equation}\label{lamda}
    A\vec{U} = \lambda\vec{U}
\end{equation}

An eigenvalue $\lambda$ of each node in the Signature-Graph is multiplied in the corresponding row pixel values, computed as node attribute $X$, as in Eq. \ref{mul}, producing the final node embeddings $\hat{X}$.
\begin{equation}\label{mul}
    \hat{X} = \Lambda X
\end{equation}
The spectral-normalised node embeddings are concatenated together in one Signature-Graph vector. This vector is concatenated with the CNN feature vector to achieve the augmentation objective. The final augmented vector is passed to a fully-connected layer.

Finally, the SGN concatenates the Signature-Graph vector $\hat{X}$ and the CNN vector $Conv$.
\begin{equation}
    f(I) = \hat{X} \frown Conv
\end{equation}
where $\frown$ denotes the concatenation process. We then apply the $Cross-Entropy$ loss function that combines the Log Softmax and negative log likelihood loss.
\begin{equation}
    Cross-Entropy = - \sum^n_{i=1}\sum^m_{j=1} y_{i,j}log(\hat{y}_{i,j})
\end{equation}
where $y_{i,j}$ refers to ground-truth label of sample $i$ belongs class $j$, and $\hat{y}_{i,j}$ is the SGN model prediction.

\subsection{An Extension to Multi-Head Attention}
SGN can be easily extended with other representation learning techniques, such as multi-head attention. 
% \textcolor{red}
{The multi-head self-attention mechanism \cite{vaswani2017attention} learns the representation encoding into through a set of key-values pairs while both the keys $\mathbf{K}$ and values $\mathbf{V}$ are the encoder hidden states. It then map the encoded pairs of key and values vectors with query $\mathbf{Q}$ vectors.}

The output dot-product attention is computed as a weighted sum of the values. The weight of each value is calculated by the dot-product of the query with all the keys, as follows:

\begin{equation}
\operatorname{Attention}(\mathbf{Q}, \mathbf{K}, \mathbf{V})=\operatorname{softmax}\left(\frac{\mathbf{Q} \mathbf{K}^{\top}}{\sqrt{n}}\right) \mathbf{V}
\end{equation}

\begin{figure*}[!ht]
\begin{center}
   \includegraphics[width=0.8\linewidth]{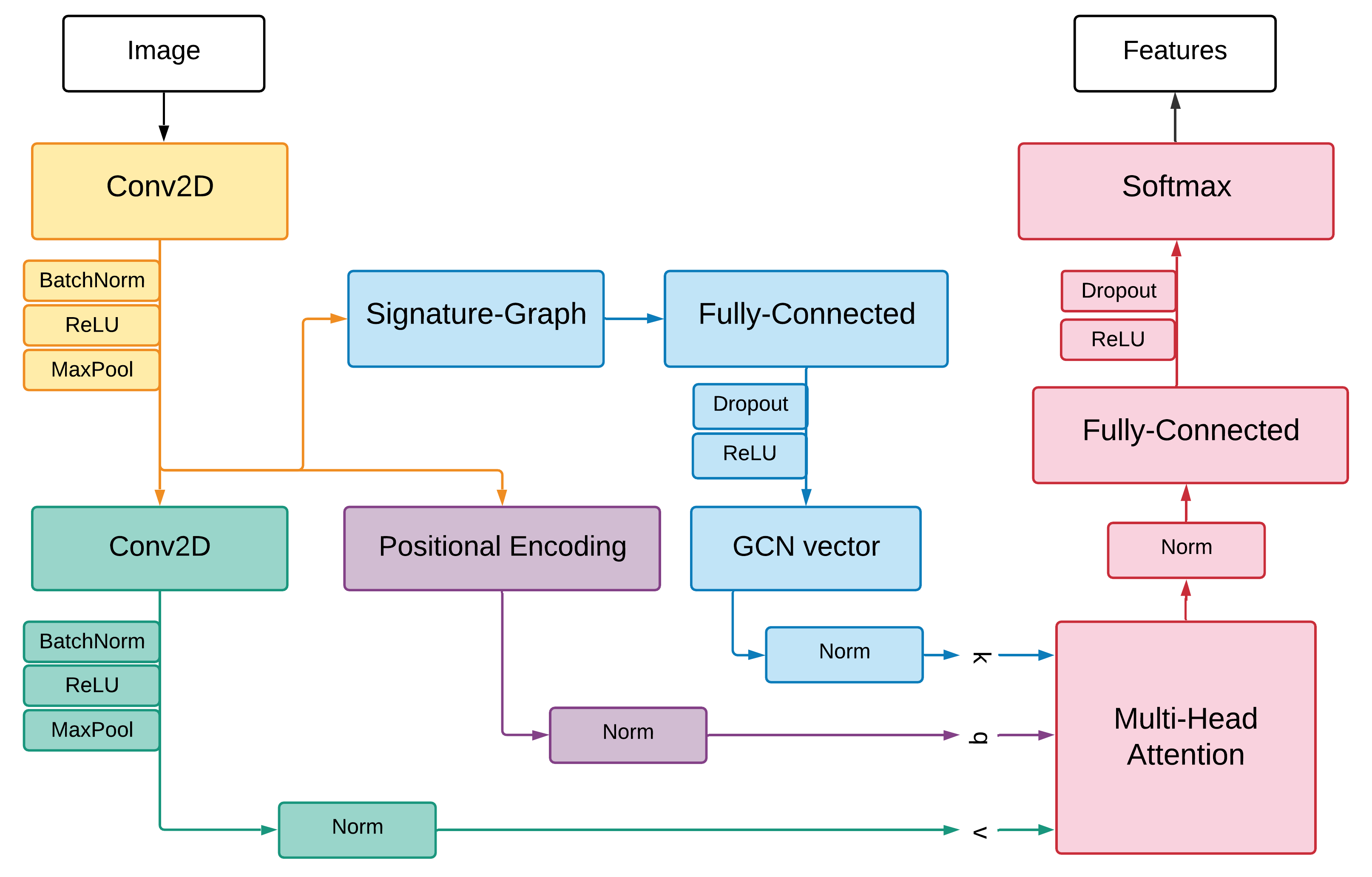}
\end{center}
   \caption{Adding an SGN layer (blue coloured) to the implementation of the head model of Multi-Head Attention.} 
   \label{fig:SGN_MHA}
\end{figure*}

We propose a novel mechanism to produce the multi-head attention key, value, and query vectors, as in Fig. \ref{fig:SGN_MHA}. the proposed SGN is utilised to encode the key vectors from the convolution maps. The convolutional feature vectors are used as the values. Positional encoding is applied on the convolutional feature to compute the query vectors. 

\begin{equation}
\begin{array}{r}
\operatorname{MHA}(Q, K, V)=\operatorname{Concat}\left(h e a d_{1}, \ldots, \text { head }_{h}\right) W^{O} \\
\text {head }_{i}=\text { Attention }\left(Q W_{i}^{Q}, K W_{i}^{K}, V W_{i}^{V}\right)
\end{array}
\end{equation}

\section{Experimental Results}
We present extensive performance analyses on seven benchmark dataset, including ASIRRA, MNIST, Fashion-MNIST, CIFAR-10, CIFAR-100, STL-10, and SVHN.
Fig. \ref{VAGCN_2} shows the neural network end-to-end trainable architecture of SGN. We implement SGN with only three convolutional blocks, each of which has followed by batch normalisation, Relu activation, and max-pooling layers. 
% CNN
The experimental results prove that this shallow architecture can outperform the state-of-the-art deep architecture on the ASIRRA data. The other datasets have lower resolution therefor we employ ResNet-50 as base model to SGN. 
% CNN
SGN is also able to beat the recent attentional and variational auto-encoders. We have also extended the SGN with a multi-head attention as designed in Fig. \ref{fig:SGN_MHA}.

\subsection{Benchmark Experiment Results}

\subsubsection{ASIRRA}
We experimented the proposed SGN on the ASIRRA (Animal Species Image Recognition for Restricting Access) dataset. Microsoft publishes this dataset for binary classification. ASIRRA is a balanced dataset having, for each class, $12,500$ and $1,000$ images for training and testing, respectively.
Table \ref{ASIRRA } shows the experimental results using the binary image classification dataset ASIRRA . 
% \textcolor{red}
{Table \ref{ASIRRA } shows the best result of using SGN with $1$ layers, $horizontal$ design, and local $maxima$.} 
SGN outperforms state-of-the-art networks with 99.65\%. flexgrid2vec \cite{hamdi2020flexgrid2vec} comes in the second rank with 98.8\%. Comparison to flexgrid2vec is important because it implements GCN over image patches. These results support the significance of the utilisation of the graph-based feature learning for image classification against the other deep CNN architectures. Besides, the table lists the accuracy scores by multiple state-of-the-art deep networks such VGG \cite{simonyan2014very}, Inception \cite{szegedy2016rethinking}, DenseNet \cite{huang2017densely}, and NASNet \cite{zoph2018learning}. The proposed SGN outperforms all of these deep networks. 

\begin{table}[!ht]
\begin{center}
\caption{Classification accuracy (top 1) results on ASIRRA  data.}\label{ASIRRA }
\small
\begin{tabular}{lc}
\toprule
Model & Test Accuracy  \\ 
\midrule
VGG16 \cite{simonyan2014very} & 82.2\% \\ %\hline
VGG19 \cite{simonyan2014very} & 88.2\% \\ %\hline
InceptionV3 \cite{szegedy2016rethinking} & 97.2\% \\ %\hline
DenseNet121 \cite{huang2017densely} & 94.3\% \\ %\hline % CVPR 2017
MobileNet \cite{howard2017mobilenets} & 98.4\% \\ %\hline
NASNetMobile \cite{zoph2018learning} & 97.9\% \\ %\hline
flexgrid2vec \cite{hamdi2020flexgrid2vec} & 98.8\% \\ %\hline
\midrule
\textbf{SGN} & \textbf{99.65\%} \\ \bottomrule
\end{tabular}
\end{center}
\end{table}%\vspace{-5mm}
   
\subsubsection{MNIST}
MNIST (Modified National Institute of Standards and Technology) dataset is composed of handwritten digits. It has $60,000$ images for training and $10,000$ for testing. The images are all in the size of $28 \times 28$ pixels. This dataset has been widely utilised to evaluate the visual pattern recognition models. 
Table \ref{MNIST} lists the benchmark results of the state-of-the-art methods using the MNIST dataset. 
% \textcolor{red}
{We have experimented SGN with various layers, design, and node locations. In Table \ref{MNIST}, we report the best result of using SGN with $1$ layers, $horizontal$ edge connection, and local $maxima$.} 
The proposed SGN outperforms the existing techniques and achieves state-of-the-art accuracy of 99.91\%. The MNIST image classification error is reduced from $0.4$ to $0.09$. Although the current methods reach high performances on the MNIST dataset, SGN increases the accuracy yet higher. 
SGN outperforms other methods with different techniques such as stochastic optimisation of CNN \cite{assiri2020stochastic}, branching and merging CNN with homogeneous filter capsules \cite{byerly2020branching}, unsupervised learning of feature representation \cite{Wang2016C-SVDDNet}.
% , and random multi-model deep learning for classification \cite{kowsari2018rmdl}. 

\begin{table}[!ht]
\begin{center}
\caption{Classification accuracy (top 1) results on MNIST.}\label{binary}\label{MNIST}
\small
\begin{tabular}{lc}
\toprule
Model & Test Accuracy  \\ 
\midrule
% Weighted Tsetlin Machine \cite{phoulady2019weighted} & 98.5\% \\ \hline
% Tsetlin Machine \cite{granmo2018tsetlin} & 98.2\% & 1.8\% \\ \hline
ProjectionNet \cite{ravi2017projectionnet} & 95\% \\ %\hline % Google research
ANODE \cite{NIPS2019_8577} & 98.2\%  \\ %\hline % NeurIPS 2019
Novikov et al., \cite{NIPS2015_5787} & 98.2\% \\ %\hline % NeurIPS 2015
Baikal \cite{gonzalez2020improved} & 99.47\%  \\ %\hline % CEC 2020
Neupde \cite{sun2020neupde} & 99.49\% \\ %\hline % PMLR 2020
TextCaps \cite{Jayasundara_2019} & 99.71\%  \\ %\hline % WACV 2019
SpinalNet \cite{kabir2020spinalnet} & 99.72\% \\ %\hline
LocalLearning \cite{nokland2019training} & 99.74\%  \\ %\hline % ICML 2019
CapsNet \cite{sabour2017dynamic} & 99.75\%  \\ %\hline % NeurIPS 2017 
% RMDL (30 RDLs) \cite{kowsari2018rmdl} & 99.82\% \\ \hline
SOPCNN \cite{assiri2020stochastic} & 99.83\% \\ %\hline % MLDM 2019
Branching/Merging CNN \cite{byerly2020branching} & 99.84\% \\ %\hline
C-SVDDNet \cite{Wang2016C-SVDDNet} & 99.6\%  \\%\hline % Pattern Recognition 2016
\midrule
\textbf{SGN} &
\textbf{99.91\%} \\ \bottomrule

\end{tabular}
\end{center}
\end{table}

\subsubsection{Fashion-MNIST}
The Fashion-MNIST \cite{xiao2017FashionMNIST} dataset has been recently utilised to evaluate the image classification models. Fashion-MNIST has $50,000$ images for training and $10,000$ for testing. Each image is a $28 \times 28$ grayscale. Fashion-MNIST covers $10$ classes of different clothes such as T-shirt, trousers, pullover, and dress.

{Table \ref{Fashion-MNIST} shows the experimental results using the Fashion-MNIST dataset. Although the Fashion-MNIST dataset is more challenging that the MNIST the proposed SGN significantly outperforms the state-of-the-art methods. 
% \textcolor{red}
{Table \ref{Fashion-MNIST} lists the best result of using SGN with $1$ layers, $vertical$ design, and local $maxima$. 
SGN achieve the highest accuracy of $98.55\%$.}
%, followed by Fine-Tuning DARTS \cite{tanveer2020fine} with $96.91\%$. 
SpinalNet \cite{kabir2020spinalnet} has $85.98\%$ and $86.61\%$ with different configurations. Neupde \cite{sun2020neupde} also has different results of $88.33\%$ and $92.40\%$ with around $5\%$ less accuracy than the proposed SGN. We present SGN as a vector-augmentation technique that complements the efforts of data augmentation methods.  Random erasing data augmentation \cite{zhong2020random} utilised different, achieving $96.35\%$ as its highest accuracy. The random erasing method employed other image classification deep architecture such as ResNet and ResNeXt by which random erasing achieve $95.99\%$ and $96.21\%$, respectively. Using only three convolutional layers with the proposed Signature-Graph method achieves the best accuracy having a new state-of-the-art result.}

\begin{table}[!ht]
\begin{center}
\caption{Classification accuracy (top 1) results on Fashion-MNIST.}\label{Fashion-MNIST}
\small
\begin{tabular}{lc}
\toprule
Model & Test Accuracy  \\ 
\midrule
% Spinal FC (8) \cite{kabir2020spinalnet} & 85.98\% & 14.02\% \\ \hline
Spinal FC (10) \cite{kabir2020spinalnet} & 86.61\% \\ %\hline
MLP 256-128-100 \cite{sun2020neupde} & 88.33\% \\ %\hline % PMLR 2020
Xiao et al., \cite{xiao2017fashion} & 89.70\%  \\ %\hline
Shake-Shake \cite{foret2021sharpnessaware} & 91.4\%  \\ %\hline % ICML 2021
% Tsetlin Machine \cite{granmo2018tsetlin} & 91.4\%  \\ \hline
ResNet18 \cite{he2016deep} & 92.00\%   \\ %\hline % CVPR 2016
Neupde \cite{sun2020neupde} & 92.40\%   \\ %\hline % PMLR 2020
% Bhatnagar et al., \cite{bhatnagar2017classification} & 92.54\% &  7.56\% \\ \hline
TextCaps \cite{Jayasundara_2019} & 93.71\%  \\ %\hline % WACV 2019
DeepCaps \cite{rajasegaran2019deepcaps} & 94.46\%  \\ %\hline % CVPR 2019
SpinalNet \cite{kabir2020spinalnet} & 94.68\% \\ %\hline
LocalLearning \cite{nokland2019training} & 95.47\%  \\ %\hline % ICML 2019
E2E-3M \cite{phong2020rethinking} & 95.92\%  \\ %\hline
ResNet + RE \cite{zhong2020random} & 95.99\%  \\ %\hline % AAAI 2020
% ResNeXt + RE \cite{zhong2020random} & 96.21\% & 3.79\% \\ \hline % AAAI 2020
WRN + RE \cite{zhong2020random} & 96.35\% \\ %\hline % AAAI 2020
Random Erasing (RE) \cite{zhong2020random} & 96.35\% \\ %\hline % AAAI 2020
% PreAct-ResNet18 + FMix \cite{} & 96.36\% & 3.64\% \\ \hline
% DARTS(2nd order) + cutout + RE \cite{liu2018darts} & 96.57\% & 3.43\% \\ \hline
% Fine-Tuning DARTS \cite{tanveer2020fine} & 96.91\% & 3.09\% \\ \hline
\midrule
\textbf{SGN} & \textbf{98.55}\% \\ \bottomrule
\end{tabular}
\end{center}
\end{table}%\vspace{-5mm}

\subsubsection{CIFAR-10 and CIFAR-100}
We also test the proposed SGN on the CIFAR-10 \cite{krizhevsky2009learning} dataset. CIFAR-10 consists of $60,000$ images. The images are in size $32 \times 32$ and divided into $50,000$ for training and $10,000$ for testing. CIFAR-10 has $10$ categories such as air-plane, automobile, bird, cat, deer, dog, frog. The categories are mutually exclusive with no overlapping. The CIFAR-100 is similar to the CIFAR-10, while having $100$ classes containing 600 images each.

Table \ref{CIFAR-10} lists the benchmark results using the CIFAR-10 dataset and the state-of-the-art methods.
% \textcolor{red}
{Although we have experimented SGN with various parameter values, Table \ref{CIFAR-10} reports the best result of using SGN with $1$ layers, $horizontal$ design, and local $maxima$ for node localisation. }
The proposed SGN outperforms the existing methods achieving $95.26\%$. 
% SGN has the best accuracy over deep architectures such as the VGG \cite{simonyan2014very}, and ResNet \cite{he2016deep}, and DenseNet \cite{huang2017densely}. 
SGN can be described as a CNN with a patch-based Signature-Graph. DeepInfoMax \cite{hjelm2018learning} is one the recent patch-based visual representation learning. As can be seen in Table \ref{CIFAR-10}, DeepInfoMax has $52.84\%$, $70.60\%$, $73.62\%$, and $75.57\%$ accuracy based on different configurations and loss functions. 
% The proposed SGN has managed to achieve better accuracy of 95.26\% with 15.23\%  accuracy more than the best results from the DeepInfoMax \cite{hjelm2018learning}. 
Moreover, SGN outperforms other state-of-the-art feature learning methods such as AutoEncoder (AE), Adversarial AE, Variational AE, and $\beta$-VAE those have achieved $55.78\%$, $57.19\%$, $57.89\%$, and $60.54\%$, respectively. 
% SGN produces at least 29.60\% higher image classification on the CIFAR-10 dataset. 
SGN also has better accuracy than recent studies such as Baikal \cite{gonzalez2020improved} with $84.53\%$, Generic Feature Extractors (GFE) \cite{Hertel2015Deep} with $89.1\%$, and CapsNet \cite{sabour2017dynamic} with $89.4\%$. Generative adversarial neural networks (GANs) have recently achieved high performances in image representation learning. However, the proposed SGN outperforms recent GANs such as BiGAN \cite{makhzani2015adversarial,dumoulin2016adversarially} with $62.74\%$ and DCGAN \cite{radford2015unsupervised} with $82.8\%$. 
APAC \cite{sato2015apac} is another work based on data augmentation. It proposes a methodology of learning augmented data. SGN has better accuracy with the proposed vector-augmentation technique.
SGN has also outperformed recent work such as MIM \cite{liao2016importance}, CLS-GAN \cite{qi2020loss}, DSN \cite{lee2015deeply}, and BinaryConnect \cite{courbariaux2015binaryconnect}.
On the CIFAR-100, SGN has achieved $84.71\%$ outperforming recent studies such as MixMatch \cite{NEURIPS2019_MixMatch}
Mish \cite{misra2020mish}, DIANet \cite{huang2020dianet}, and ResNet-1001 \cite{he2016identity}. Table \ref{CIFAR-100} shows more experimental results on the CIFAR-100. We have experimented SGN with various settings. 
% \textcolor{red}
{In Table \ref{CIFAR-100}, SGN has the best results with $1$ layers, $horizontal$ design, and local $maxima$.} 

\begin{table}[!ht]
\begin{center}
\caption{Classification accuracy (top 1) results on CIFAR-10.}\label{CIFAR-10}
\small
\begin{tabular}{lc}
\toprule
Model & Test Accuracy  \\ 
\midrule
% Noise As Target (NAT) \cite{bojanowski2017unsupervised} & 51.29\%\\ \hline
% DeepInfoMax (G) \cite{hjelm2018learning} & 52.84\%\\ \hline
AutoEncoder (AE) & 55.78\%\\ %\hline
Adversarial AE \cite{makhzani2015adversarial} & 57.19\%\\ %\hline
$\beta$-VAE \cite{higgins2016beta,alemi2016deep} & 57.89\%\\ %\hline
% Variational (VAE) \cite{kingma2013auto} & 60.54\%\\ \hline
ANODE \cite{NIPS2019_8577} & 60.6\% \\ %\hline % NeurIPS 2019
BiGAN \cite{makhzani2015adversarial,dumoulin2016adversarially} & 62.74\%\\ %\hline
% DeepInfoMax (DV) \cite{hjelm2018learning} & 70.60\%\\ \hline % ICLR 2019
% DeepInfoMax (JSD) \cite{hjelm2018learning} & 73.62\%\\ \hline % ICLR 2019
DeepInfoMax (infoNCE) \cite{hjelm2018learning} & 75.57\%\\ %\hline % ICLR 2019
DenseNet \cite{huang2017densely} & 77.79\% \\ %\hline % CVPR 2017
% DenseNet (Spinal FC) & 81.13\%\\ \hline
DCGAN \cite{radford2015unsupervised} & 82.8\% \\ %\hline
% Roto-Scat + SVM \cite{oyallon2015deep} & 82.30\% \\ \hline
% ExemplarCNN \cite{dosovitskiy2014discriminative} & 84.3\% \\ \hline
% Baikal \cite{gonzalez2020improved} & 84.53\% \\ %\hline % CEC 2020
Scat + FC \cite{oyallon2017scaling} & 84.7\% \\ %\hline
% VGG-11 \cite{simonyan2014very} & 86.68\%\\ \hline
% VGG-11 (Spinal FC) & 87.08\%\\ \hline
% VGG-13 \cite{simonyan2014very} & 87.79\%\\ \hline
FPID \cite{pmlr-v80-hoffman18a} & 89.06\% \\ %\hline
GFE \cite{Hertel2015Deep} & 89.1\% \\ %\hline % Generic Feature Extractors IJCNN'15
CapsNet \cite{sabour2017dynamic} & 89.4\% \\ %\hline % NeurIPS 2017 
MP \cite{hendrycks2016baseline} & 89.07\%\\ %\hline
VGG-13 (Spinal FC) & 89.16\% \\ %\hline
ResNet-34 \cite{he2016deep} & 89.56\% \\ %\hline
APAC \cite{sato2015apac} & 89.70\% \\ %\hline

MIM \cite{liao2016importance} & 91.5\% \\ %\hline % WACV 2016
CLS-GAN \cite{qi2020loss} & 91.7\% \\ %\hline % IJCV'2020
DSN \cite{lee2015deeply} & 91.8\% \\ %\hline % PMLR'15
BinaryConnect \cite{courbariaux2015binaryconnect} & 91.7\% \\ %\hline % NeurIPS 2015
Mish \cite{misra2020mish} & 92.20\% \\ %\hline % BMVC 2020
\midrule
\textbf{SGN} & \textbf{95.26\%} \\ \bottomrule
\end{tabular}
\end{center}
\end{table}%\vspace{-5mm}

\begin{table}[!ht]
\begin{center}
\caption{Classification accuracy (top 1) results on CIFAR-100.}\label{CIFAR-100}
\small
\begin{tabular}{lc}
\toprule
Model & Test Accuracy  \\ 
\midrule
DSN \cite{lee2015deeply} & 65.4\% \\ %\hline % PMLR'15 
Unsharp Masking \cite{carranza2019unsharp} & 60.36\% \\ %\hline % ICANN 2019  
ResNet-50  \cite{he2016identity} & 67.06\% \\ %\hline % PMLR'15 
GFE \cite{Hertel2015Deep} & 67.7\% \\ %\hline % Generic Feature Extractors IJCNN'15
MIM \cite{liao2016importance} & 70.8\% \\ %\hline % WACV 2016
MixMatch \cite{NEURIPS2019_MixMatch} & 74.10\% \\ %\hline % NIPS 2019 
Mish \cite{misra2020mish} & 74.41\% \\ %\hline % BMVC 2020 
Stochastic Depth \cite{huang2016deep} & 75.42\% \\ %\hline % ECCV 2016  Cited by 1102
Exponential Linear Units \cite{clevert2015fast} & 75.7\% \\ %\hline % ArXiv Cited by 3233
DIANet \cite{huang2020dianet} & 76.98\% \\ %\hline %  AAAI2020
Evolution \cite{real2017large} & 77\% \\ %\hline %  ICLR 2017
ResNet-1001 \cite{he2016identity} & 77.3\% \\ %\hline %  ECCV2016 Cited by 5K+
\midrule
\textbf{SGN} & \textbf{84.71\%} \\ \bottomrule
\end{tabular}
\end{center}
\end{table}%\vspace{-1mm}

\subsubsection{STL-10}
The STL-10 \cite{coates2011analysis} is prepared for image recognition model evaluation. STL-10 has fewer labelled training examples. It has $5,000$ images for training and $8,000$ for testing, over $10$ classes. The STL-10 has $100,000$ unlabelled images for unsupervised learning. However, in this paper, we did not use any of these unlabelled images. The STL-10 $5,000$ and $8,000$ images are in the size of $96 \times 96$ and are acquired for the ImageNet. 
% \textcolor{red}
{Table \ref{STL-10} lists the evaluation results using the STL-10 dataset using SGN with $1$ layers, $horizontal$ edge connections, and node located on local $maxima$.} 
SGN outperforms the state-of-the-art methods. Consistently with the benchmark results on CIFAR-10, SGN has managed to outperform different methodologies such as the patch-based DeepInfoMax \cite{hjelm2018learning}, Autoencoders \cite{makhzani2015adversarial,kingma2013auto,higgins2016beta,alemi2016deep,dumoulin2016adversarially}, and GAN \cite{makhzani2015adversarial, dumoulin2016adversarially}. The DeepInfoMax has scored $28.09\%$, $61.92\%$, $65.93\%$, and $67.08\%$. SGN outperforms all of these results with $70.94\%$ accuracy.

\begin{table}[!ht]
\caption{Classification accuracy (top 1) results on STL-10 dataset.}\label{STL-10}
\centering
\begin{tabular}{lc}
\toprule
Model & Test Accuracy  \\ 
\midrule
% DeepInfoMax (G) \cite{hjelm2018learning} &  28.09\% \\ \hline
Adversarial AE \cite{makhzani2015adversarial} & 43.89\%\\ %\hline
Variational AE \cite{kingma2013auto} & 46.47\%\\ %\hline
AE & 46.82\%\\ %\hline
$\beta$-VAE \cite{higgins2016beta,alemi2016deep} & 46.87\%\\ %\hline
% Noise As Target (NAT) \cite{bojanowski2017unsupervised} & 48.84\% \\ \hline
BiGAN \cite{makhzani2015adversarial,dumoulin2016adversarially} & 58.48\% \\ %\hline
% Convolutional K-means \cite{coates2011selecting} & 60.1\% \\ \hline
DeepInfoMax (DV) \cite{hjelm2018learning} & 61.92\% \\ %\hline % ICLR 2019
% NN-Weighter \cite{ericsson2020don} & 63.13 \\ \hline
% Accuracy Monitoring \cite{shao2020increasing} & 63.43\%\\ \hline
% Hierarchical Matching Pursuit (HMP)\cite{oyallon2017scaling} & 64.5\% \\ \hline
DeepInfoMax (JSD) \cite{hjelm2018learning} & 65.93\% \\ %\hline % ICLR 2019
DeepInfoMax (infoNCE) \cite{hjelm2018learning} & 67.08\% \\ %\hline % ICLR 2019
% DFF Committees \cite{Miclut_2014} & 68\%\\ \hline
% RotNet \cite{ericsson2020don} & 68.19 \\ \hline
% C-SVDDNet \cite{Wang2016C-SVDDNet} & 68.20\%\\ \hline
% L2RW \cite{ericsson2020don} & 69.15\% \\ \hline
% BDW \cite{ericsson2020don} & 71.12\% \\ \hline 
ResNet18 \cite{luo2020extended} & 72.66\% \\ %\hline
Second-order Hyperbolic CNN \cite{ruthotto2019deep} & 74.3\% \\ %\hline % MIV 2019 (Q1)
% SWWAE \cite{zhao2015stacked} &  74.30\%	\\ \hline % Google Research
SOPCNN (RA) \cite{NEURIPS20_FixMatch} &  88.08\%	\\ %\hline % NEURIPS 2020
FixMatch  \cite{NEURIPS20_FixMatch} &  89.59\%	\\ %\hline % NEURIPS20
SESN \cite{sosnovik2019scale} &  91.49\%	\\ %\hline % iclr'19
NSGANetV2 \cite{lu2020nsganetv2} &  92\%	\\ %\hline % eccv'20
FixMatch (RA) \cite{NEURIPS20_FixMatch} &  92.02\%	\\ \hline % NEURIPS 2020
\textbf{SGN} & \textbf{94.36\%}\\ \bottomrule
\end{tabular}
\end{table}%\vspace{-2mm}

\subsubsection{SVHN}
We have utilised the street view house numbers (SVHN) dataset to evaluate the proposed SGN. The SVHN contains $630,420$ house numbers images. Table \ref{SVHN} lists the benchmark results on the cropped format of the SVHN with $32 \times 32$ pixels.
The SVHN official split contains $73,257$ and $26,032$ image for training and testing, respectively. 
% \textcolor{red}
{SGN has the best accuracy of $95.86$\% when using $1$ layers, $vertical$ design, and local $minima$.} SGN outperforms existing methods such as RetNet, DANN, and DWT. 

\begin{table}[!ht]
\begin{center}
\caption{Classification accuracy (top 1) results on SVHN data.}\label{SVHN}
\small
\begin{tabular}{lc}
\toprule
Model & Test Accuracy  \\ 
\midrule
Asymmetric Tri-Training  \cite{saito2017asymmetric} & 90.83\% \\ %\hline
ReNet+LSTM \cite{Moser2020DartsReNet} & 94.10\% \\ %\hline
DenseNet \cite{huang2017densely} & 94.19\% \\ %\hline % CVPR 2017
WRN-OE \cite{hendrycks2019deep} & 94.19\% \\ %\hline
WRN \cite{zagoruyko2016wide}    & 94.50\% \\ %\hline
ReNet+GRU \cite{Moser2020DartsReNet} & 95.16\% \\ %\hline
Farhadi et al., \cite{farhadi2019novel} & 94.62\% \\ %\hline
FPID \cite{pmlr-v80-hoffman18a} & 95.67\% \\ %\hline
E-ABS \cite{ju2020abs} & 89.20\% \\ %\hline
DANN \cite{ganin2016domain} & 91.00\% \\ %\hline
Associative Domain Adaptation \cite{haeusser2017associative} & 91.80\% \\ %\hline
SE-a \cite{french2017self} & 91.92\% \\ %\hline
SE-b \cite{french2017self} & 95.62\% \\ %\hline
CLS-GAN \cite{qi2020loss} & 94.02\% \\ %\hline  % IJCV'2020
DWT-MEC \cite{roy2019unsupervised} & 94.62\% \\ %\hline 
\midrule
\textbf{SGN} & \textbf{95.86\%} \\ \bottomrule
\end{tabular}
\end{center}
\end{table}%\vspace{-5mm}

\subsection{Ablation Study}
SGN has been implemented with two versions of network architecture with different number of convolutional blocks and edge-connection design. We have changed the network architecture design based on four criteria, as follows:
\begin{itemize}
    \item Node coordinates of local maxima and minima of the CNN features in the image patches.
    \item Visual feature embeddings via horizontal and vertical Signature-Graphs.
    \item The number of the SGN layers is experimented on of one-, two-, or three-layer.
    \item Using Signature-Graph layers with and without skip connection.
    \item Using SGN with pre-trained based models as well as the multi-head attentional mechanism.
    \item Patch size is experimented on $6\times 6$ and $10\times 10$.
\end{itemize}%\vspace{-5mm}

\begin{table*}[]
\begin{center}
\caption{Using SGN with local maxima against minima on the Fashion-MNIST and CIFAR-10.}\label{ablation2}
\small
\begin{tabular}{ccccc}
\toprule
Layers & Edege-connection Mode &  Fashion-MNIST (Maxima) & Fashion-MNIST (Minima) &   CIFAR-10 (Maxima)\\ %\hline
\midrule
1      & Horizontal & \textbf{98.55}\% & 93.24\% & \textbf{95.26\%} \\%\hline
2      & Horizontal & 93.04\% & 92.91\% & 84.27\% \\%\hline
3      & Horizontal & 92.38\% & 92.78\% & 86.26\%  \\%\hline
1      & Vertical   & 97.08\% & 92.86\% & 89.13\% \\%\hline
2      & Vertical   & 94.82\% & 91.90\% & 88.05\% \\%\hline
3      & Vertical   & 95.11\% & 93.01\% & 87.70\% \\%\hline
\bottomrule
\end{tabular}
\end{center}
\end{table*}

\begin{figure*}[]
\begin{center}
   \includegraphics[width=0.88\linewidth]{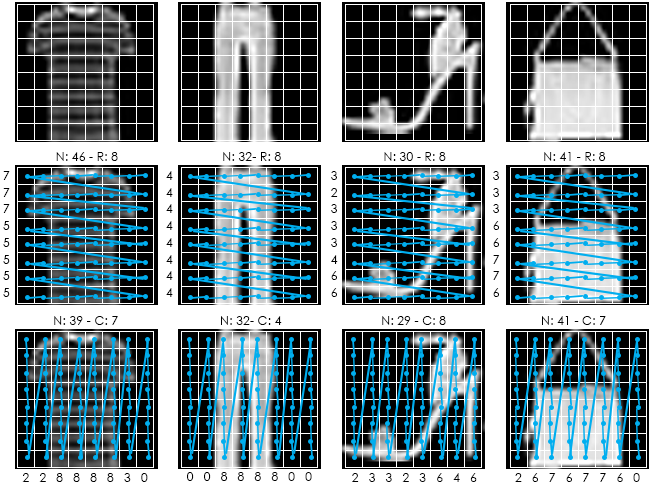}
\end{center}
   \caption{Using Signature-Graph with horizontal and vertical edge-connection modes on four samples form the Fashion-MNIST dataset. The first row shows original images sliced into equal, non-overlapping patches. The second and third rows visualise the horizontal and vertical edge connection designs of Signature-Graph. We added the statistics on how many non-empty rows (R) and columns (C) above each image. We have also calculated the number of non-empty patches in each image to the effectiveness of each edge connection design.}
   \label{SGN-FM}
\end{figure*}
%

% \textcolor{red}
\subsubsection{Node Embedding and Edge Connection:}
Table \ref{ablation2} compares the SGN performance according to different node-embedding based on local maxima and minima of the CNN features. Using horizontal mode over local maxima-based nodes has the best results with $98.55$\% and $95.26$\%  in the Fashion-MNIST and CIFAR-10 datasets, respectively. Signature-Graph designed on local maxima, and horizontal edge-connection outperformed other ablations. Fig. \ref{SGN-FM} has three rows where the first row shows four sample images from the Fashion-MNIST dataset, the second row visualises the horizontal Signature-Graph, and the third one visualises the vertical design. The pixel distribution over rows tends to be better than columns. Usually, every row has non-empty regions. In contrast, many columns have no foreground values. These facts support the superiority of horizontal SGN over vertical design. Fig. \ref{SGN-FM} shows counts of non-empty image patches in each row (horizontal) and column (vertical) designs. The totals of non-empty nodes and full rows or columns are given on top of each image. For example, the t-shirt image has higher values of nodes and rows using the horizontal Signature-Graph than the vertical one. These numbers of nodes (patches) and horizontal rows or vertical columns directly impact the feature space. 

\subsubsection{SGN Layers and Skip Connection}
\begin{figure}[]
\begin{center}
   \includegraphics[width=0.9\linewidth]{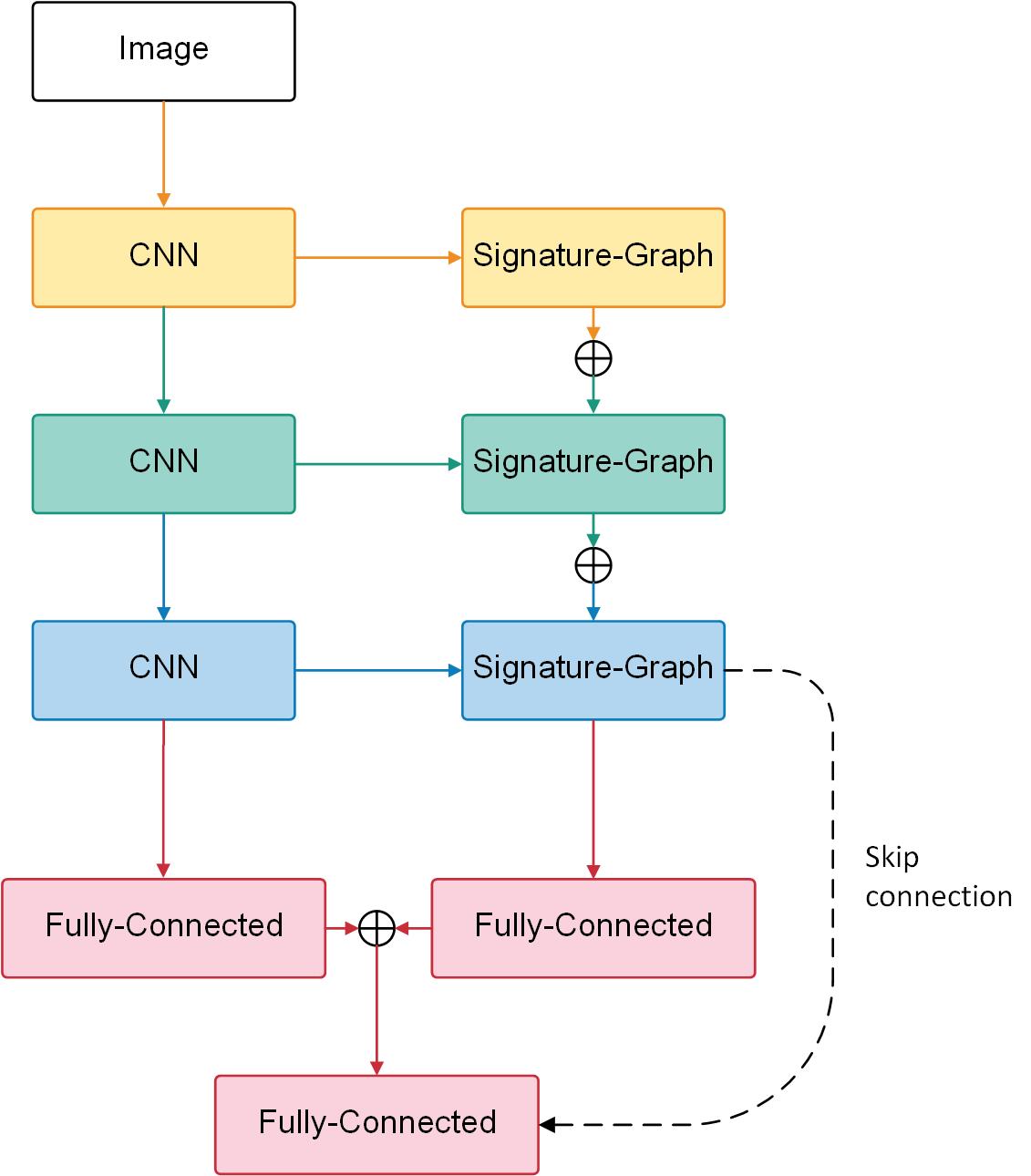}
\end{center}
   \caption{Signature-Graph with skip connection.}%\vspace{-5mm}
   \label{SGN-SC}
\end{figure}

We also tested the proposed SGN with one, two and three Signature-Graph layers. Table \ref{ablation2} shows that using one layer of the proposed horizontal Signature-Graph has the best results over one or three layers on the Fashion-MNIST dataset with $98.55$\% outperforming the vertical design. Using two Signature-Graph layers produces lower accuracy than the three-layer architecture. The latter one has $95.11$\% and $93.01$\% outperforming the two-layer method on the vertical SGN with maxima- and minima-based node locations, respectively. We tested SGN with skip connection as visualised in Fig. \ref{SGN-SC}. Skip connection has become a standard component of CNN based visual representation methods. The skip connection is usually applied as addition and concatenation such as in ResNet \cite{he2016identity} and DenseNet \cite{huang2017densely}. In this experiment, we feed the Signature-Graph features directly to the last fully-connected layer skipping other layers in between. This implementation adds the original Signature-Graph features to the final SGN feature vector allowing an alternative path for the gradient with back-propagation. Skip connection is helpful for CNN architecture convergence. The SGN with skip connection has $97.49$\% accuracy on the Fashion-MNIST compared to $98.55$\% without skip connection. This result outperforms all the ablations based on the vertical edge connection and minima-based node locations. However, using the Signature-Graph based on horizontal design and maxima node locations has the best accuracy without skip connection.  

\subsubsection{SGN with Multi-head Attention}
We have also tested the proposed SGN with and without base and head model extensions. Using the MHA with the pre-trained EfficientNet has $86.92$\% while using SGN with same pre-trained model has $88.56$\%.
Using MHA with ResNet50 has achieved $92.37\%$ while adding SGN achieve the best accuracy of $94.36$\% on the STL-10 dataset.

\begin{table}[]
\begin{center}
\caption{Classification accuracy (top 1) results using the SGN on STL-10 trained with base and head model extensions.}\label{ablation3}
\small
\begin{tabular}{lc}
\toprule
Basemodel   & Test Accuracy \\
\midrule
EfficientNet+MHA  & 86.92\%  \\ %\hline
EfficientNet+SGN  & 88.56\%  \\ %\hline
ResNet-50+MHA     & 92.37\%     \\ %\hline
ResNet-50+MHA+SGN & \textbf{94.36\%}   \\
\bottomrule
\end{tabular}
\end{center}
\end{table}

\subsubsection{SGN with on Different Patch Sizes}
We have also investigated the impact of changing the patch size when partitioning the CNN feature maps to extract the Signature-Graph. This size controls how many nodes in the graph. For example, if the input image has $96 \times 96$ pixels, it has $16$ and $10$ nodes for the sizes of $6$ and $10$, respectively. This variation has an impact on the performance of the SGN. For example, the SGN has $97.08\%$ and $92.65\%$ for $6\times 6$ and $10\times 10$ patch sizes, when test on the Fashion-MNIST with the vertical design and local maxima nodes. Therefore, we utilised the $6\times 6$ partitioning for all experiments to gain more useful Signature-Graph features.

\section{Conclusion}
% \textcolor{red}
{Signature-Graph Network (SGN) contributes to graph feature embedding, CNN vector augmentation, and attention computing.} SGN is a novel solution to the CNN limitation of the missing global structure information. SGN added a Signature-Graph layer on top of the convolutional block layers. 
% \textcolor{red}
{The Signature-Graph extracts accurate feature embeddings through a signature-like graph from the convolutional feature map. The Signature-Graph utilises spectral graph theory to normalise the nodes' embeddings.} We have compared the performance accuracy of the proposed SGN against a set of recent related works such as patch-based, auto-encoders, GAN, GCN, multi-head attention and data augmentation. The experimental results show that the proposed SGN end-to-end network using the Signature-Graph produces state-of-the-art image classification results. 
% \textcolor{red}
{SGN works with a straightforward, shallow architecture of few convolution layers augmented and normalised with spectral Signature-Graph layers in comparison to these complex methods.}

\section*{Acknowledgments}
Ali Hamdi is supported by RMIT Research Stipend Scholarship. This research is partially supported by Australian Research Council (ARC) Discovery Project \textit{DP190101485}.

\bibliographystyle{elsarticle-num}
\bibliography{bib}

\end{document}